\documentclass[10pt,twocolumn,letterpaper]{article}

\usepackage{cvpr}
\usepackage{times}
\usepackage{epsfig}
\usepackage{graphicx}
\usepackage{amsmath}
\usepackage{amssymb}
\usepackage{amsfonts}
\usepackage{algorithm}
\usepackage{algpseudocode}
\usepackage{graphics}
\usepackage{booktabs}
\usepackage{multirow}
\usepackage{makecell}
\usepackage{flushend}

\newcommand{\tabincell}[2]{\begin{tabular}{@{}#1@{}}#2\end{tabular}}
 
\usepackage[pagebackref=true,breaklinks=true,letterpaper=true,colorlinks,bookmarks=false]{hyperref}

\cvprfinalcopy 

\def\cvprPaperID{} 

\ifcvprfinal\fi
\begin{document}

\title{Mutual Learning Network for Multi-Source Domain Adaptation}

\author{Zhenpeng Li$^1$,\quad Zhen Zhao$^1$,\quad Yuhong Guo$^{1,2}$,\quad Haifeng Shen$^1$,\quad Jieping Ye$^1$\\
AI Tech, DiDi Chuxing, China$^1$, \qquad
Carleton University, Canada$^2$
}

\maketitle

\begin{abstract}
   Early Unsupervised Domain Adaptation (UDA) methods 
   have mostly assumed the setting of a single source domain, where all the labeled source data come from the same distribution. 
   However, in practice the labeled data can come from multiple source domains with different distributions.
In such scenarios, the single source domain adaptation methods can fail due to 
the existence of domain shifts across different source domains
and multi-source domain adaptation methods need to be designed.
In this paper, we propose a novel multi-source domain adaptation method, Mutual Learning Network for Multiple Source Domain Adaptation ({ML-MSDA}). 
Under the framework of mutual learning, 
the proposed method pairs the target domain with each single source domain to train a conditional adversarial domain adaptation network as a branch network,
while taking the pair of the combined multi-source domain and target domain to train a conditional adversarial adaptive network as the guidance network.
The multiple branch networks are aligned with the guidance network to achieve mutual learning 
by enforcing JS-divergence regularization 
over their prediction probability distributions on the corresponding target data.  
We conduct extensive experiments on multiple multi-source domain adaptation benchmark datasets.
The results show the proposed ML-MSDA method outperforms the comparison methods and achieves the state-of-the-art performance.
\end{abstract}

\section{Introduction}

Deep neural networks have produced great advances for many computer vision tasks, including classification, detection and segmentation.
Such success nevertheless depends on the availability of large amounts of labeled training data
under the standard supervised learning setting.  
However, the labels are typically expensive and time-consuming to produce through manual effort.
Domain adaptation aims to reduce the annotation cost by exploiting existing labeled data in auxiliary source domains.
As the data in source domains can be collected with different equipments or in different environments, 
they may exhibit different distributions from the target domain data.
Hence 
the main challenge of domain adaptation is to bridge the distribution divergence across domains 
and effectively transfer knowledge from the source domains 
to train prediction models in the target domain.
A widely studied domain adaptation setting is unsupervised domain adaptation (UDA),
where data in the source domain are labeled and 
data in the target domain are entirely unlabeled.

\begin{figure*}[t!]
\begin{center}
\includegraphics[width=.85\linewidth]{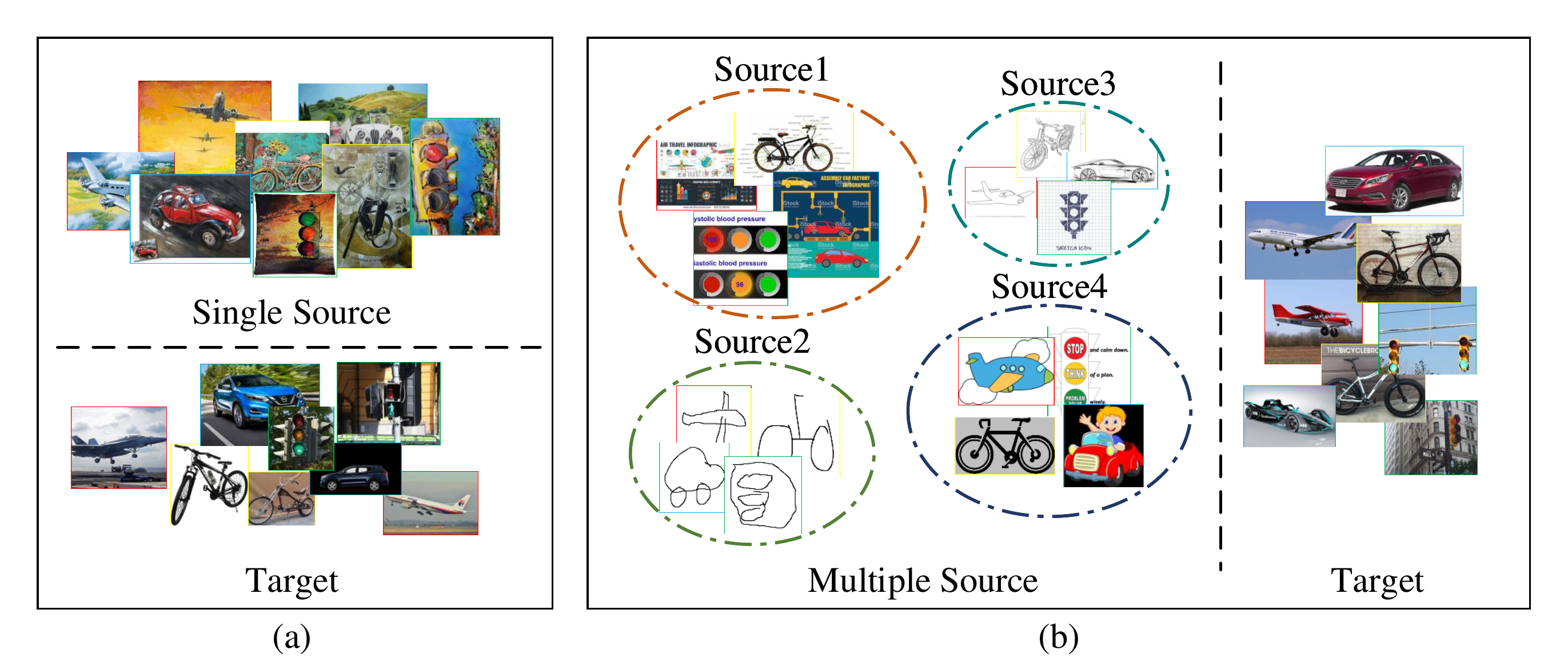}
\end{center}
\caption{(a) Single source unsupervised domain adaptation (UDA) setting, where the source domain data all come from the same distribution. 
(b) Multi-source domain adaptation (MSDA) setting, where the source data are from different domains and hence have different distributions.}
\label{fig:UDA and MSDA}
\end{figure*}

Early UDA methods assume the source domain data all come from the same source and have the same distribution, 
as shown in Figures~\ref{fig:UDA and MSDA}(a). 
In practice, it is much easier to collect labeled data from multiple source domains with different distributions,
as shown in Figures~\ref{fig:UDA and MSDA}(b).
For example, we can collect source domain data from live action movies, cartoons, hand-drawn pictures, etc. 
Exploiting data from multiple source domains has the potential capacity of transferring more useful information to the target domain, 
and can be more beneficial in practical applications.
Some recent multi-source domain adaptation (MSDA) methods 
have used shared feature extractors for different source domains 
\cite{xu2018deep,peng2019moment,zhao2018adversarial}.
The works in 
\cite{xu2018deep,peng2019moment}
make predictions in the target domain by using weighted combinations of multiple source domain results,
while the other work in 
\cite{zhao2018adversarial} trains a classifier for all source and target domains,
but back propagates only the minimum cross-domain training error among all source domains.
%
These methods however fail to handle the distribution divergence between different source domains. 
In addition, it is difficult for these methods to 
bridge gaps between the target domain and the multiple source domains simultaneously,
while negative optimization and transfer may occur~\cite{xu2018deep}. 
Therefore, how to balance the distribution difference between source-source and source-target domains is a key 
for developing effective MSDA methods.

In this paper, we propose a new approach for multi-source domain adaptation, 
namely Mutual Learning Network for Multi-Source Domain Adaptation (\textit{ML-MSDA}). 
As the multiple source domains have different distributions, 
ML-MSDA trains one separate conditional adversarial adaptation network, referred to as branch network, to align each source domain with the target domain. 
In addition, it also trains a conditional adversarial adaptation network to align the combined source domain with the target domain,
which is referred to as guidance network. 
The guidance and branch networks share weights in the first few feature extraction layers, while the remaining layers are branch specific. 
We then propose to perform guidance network centered prediction alignment 
by enforcing JS-divergence regularizations over the prediction probability distributions of target samples 
between the guidance network and each branch network so that all networks can learn from each other 
and make similar predictions in the target domain. 
Such a mutual learning structure is expected to gather 
domain specific information 
from each single source domain through branch networks 
and gather complementary common information through the guidance network,
aiming to improve both the information adaptation efficacy across domains
and the robustness of network training.
The contribution of this paper is three fold. 
First, we propose a novel mutual learning network architecture for multi-source domain adaptation,
which enables guidance network centered information sharing in the multi-source domain setting.
Second, we develop a novel dual alignment mechanism at both the feature and prediction levels: 
conditional adversarial feature alignment across
each pair of source and target domains, and centered prediction alignment between each branch network
and the guidance network. 
Third, 
we conduct experiments on multiple benchmark datasets
and demonstrate the superiority of the proposed method over the-state-of-the-art UDA and MSDA methods.


\section{Related Work}
\noindent{\bf Unsupervised Domain Adaptation with Single Source Domain.} 
Unsupervised domain adaptation (UDA) 
addresses the problem of exploiting labeled data from a source domain to
train prediction models for a target domain where all the data instances are unlabeled. 
UDA has mostly focused on the single source domain setting where the labeled source data
are collected from the same source, and hence have the same distribution. 
The key to solve UDA problems lies in eliminating or mitigating the domain shift between source and target domains. 
Many works have exploited distribution distance metrics, 
such as Maximum Mean Discrepancy (MMD) 
and Kullback-Leibler (KL) divergence, 
to reduce gaps between the statistical distributions of the source and target domains 
~\cite{long2017deep,yan2017mind,long2016unsupervised,tzeng2014deep,sun2016deep}.
Some recent works have adopted an adversarial learning based DA mechanism
\cite{ganin2016domain,liu2016coupled}, which aligns the feature distributions through a minimax adversarial game
between the feature extractor and the domain discriminator.
Following the adversarial mechanism, the networks can learn domain-invariant features across the source and target domains, 
and generate source or target data~\cite{liu2016coupled,bousmalis2017unsupervised,li2017demystifying}.
In~\cite{long2018conditional}, the authors further adopted conditional adversarial learning for unsupervised domain adaptation.
The teach-student (T-S) learning mechanism has also been used for unsupervised 
domain adaptation~\cite{french2017self,manohar2018a,li2017large-scale}.
In \cite{french2017self}, 
the teacher network is updated as
an exponential moving average of the student network,
while the prediction difference on unlabeled data
between the student and teacher networks is penalized. 
In \cite{manohar2018a,li2017large-scale}, the teacher network is trained on the source domain and the student network 
is trained on the target domain, while the teacher network is used to ``teach" the student network
on unlabeled parallel data the connect the two domains. 
\\

\noindent\textbf{Multiple Source Unsupervised Domain Adaptation.} 
In practice we can get labeled training data from multiple source domains with different distributions. 
Directly applying single domain UDA methods cannot work well in this case as they fail to address
the differences between the multiple source domains.
To address the domain shift between the multiple source domains, 
FA~\cite{daume2009frustratingly} concatenates the extra features of each source domain 
to induce properties shared between each source domain and the target domain. 
A-SVM~\cite{yang2007cross} ensembles the multiple source specific classifiers. %
The Domain Adaptation Machine~\cite{duan2012domain} integrates domain-related regularization terms 
to train a set of source classifiers and make the target classifier share similar decision values with them. 
CP-MDA~\cite{chattopadhyay2012multisource} computes weight values for the classifier of each source domain 
and uses conditional distributions to combine them. 
DCTN~\cite{xu2018deep} deploys a domain discriminator and a category classifier for each source domain and 
uses the loss of each discriminator to calculate the weight of each classifier. 
M3SDA~\cite{peng2019moment} utilizes matching moments to directly match all distributions of source and target domains. 
MDAN~\cite{zhao2018adversarial} uses adversarial adaptation to induce invariant features for all pairs of source-target domains.
Different from these related works, our proposed approach \textit{ML-MSDA}
introduces a new mutual learning network architecture that has one guidance network and multiple branch networks.
It exploits each source domain for domain adaptation in both domain specific manner (through branch networks) and 
domain ensemble manner (through the guidance network).

\begin{figure*}[th!]
\begin{center}
\includegraphics[width=0.8\linewidth]{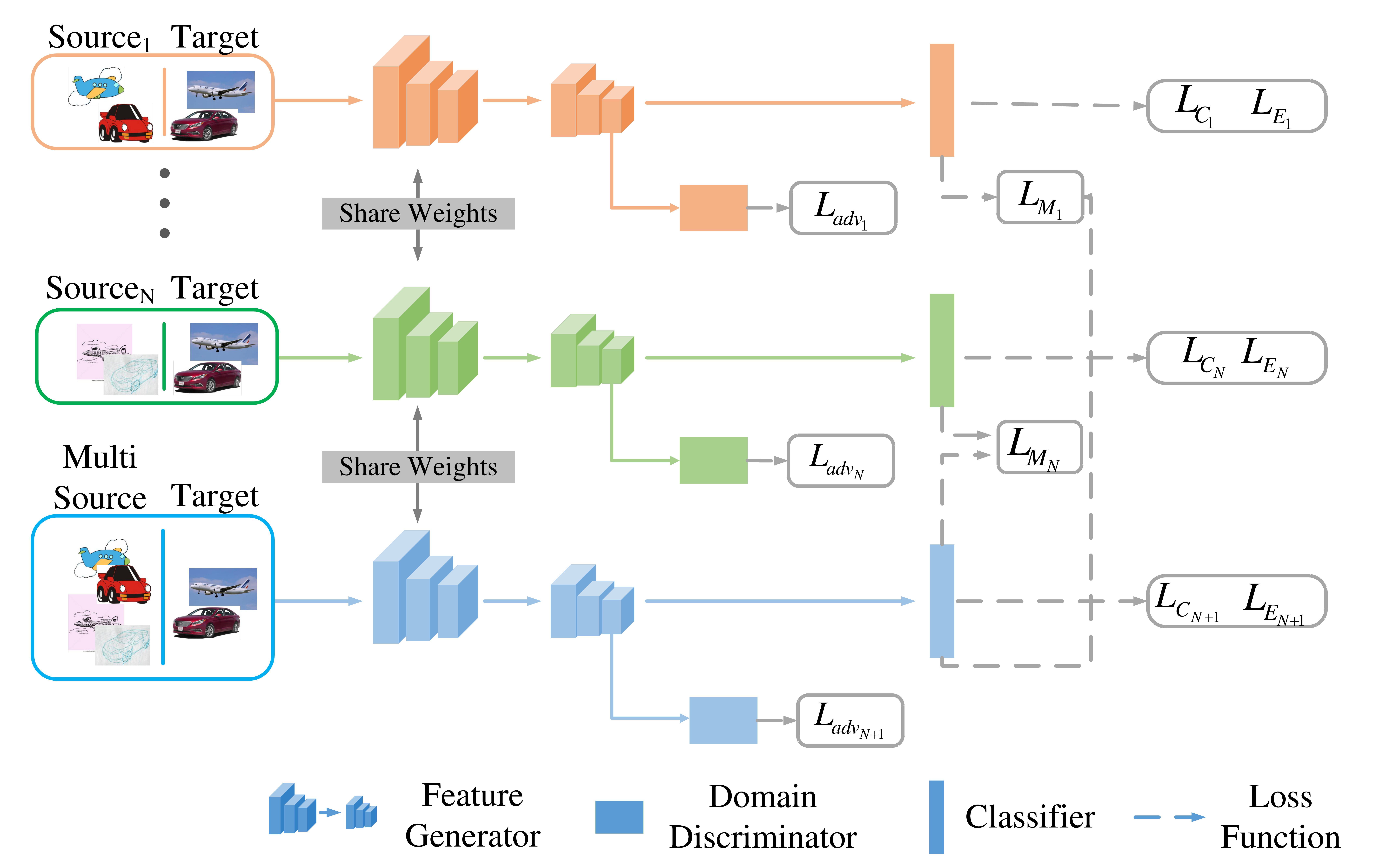}
\end{center}
	\caption{The framework of the proposed \textit{Mutual Learning network}. 
	For N source domains, it has N branch networks and one guidance network (the bottom one).  
	For each branch network, the corresponding source domain data and the target domain data are used as inputs.
	The combined multiple source domain data and the target domain data are used as inputs for the guidance network.
	All these subnetworks have the same structure that has three components: feature extractor, domain discriminator, and category classifier.  
	Classification losses, $L_C$ and $L_E$, and adversarial alignment loss $L_{adv}$ are considered on each subnetwork.
	A prediction misalignment loss $L_M$ is considered between each branch network and the guidance network.
	}
\label{fig:ML-MSDA}
\end{figure*}


\section{Mutual Learning Network for MSDA}

We consider the following multi-source domain adaptation setting.
Assume we have $N$ source domains,
\({\mathcal{D}_S} = \left\{ {\mathcal{D}_{S_j}} \right\}_{j = 1}^N\) and 
one target domain \({{\mathcal{D}_T}}\).
The multiple source domains and the target domain all have different input distributions. 
For each source domain, all the input images are labeled, such that 
\({\mathcal{D}_{{S_j}}} = \left( {{\textbf{X}_{{S_j}}},{\textbf{Y}_{{S_j}}}} \right)\)
\(=\{({\bf x}_i^j, {\bf y}_i^j)\}_{i=1}^{n_s^j}\),
where ${\bf x}_i^j$ denotes the input image and ${\bf y}_i^j\in[0,1]^K$ 
denotes the corresponding 
label indicator vector.
For the target domain, the labels of the images are unavailable,
such that \({\mathcal{D}_T} = {{\textbf{X}_T}} = \left\{ {\textbf{x}_i^t} \right\}_{i = 1}^{n_t}\).

In this section, we present a novel mutual learning network model for MSDA. 
The proposed approach is termed as Mutual Learning network for Multi-Source Domain Adaptation (ML-MSDA). 
The framework of ML-MSDA is presented in Figures~\ref{fig:ML-MSDA}.
In this learning framework, we aim to exploit both the domain specific adaptation information from each source domain
and the combined adaptation information in multiple source domains. 
We build $N+1$ subnetworks for domain adaptation. 
The first $N$ subnetworks perform domain adaptation from each corresponding single source domain to the target domain, 
while the $(N+1)$-th subnetwork performs domain adaptation from the combined multiple source domains to the target domain. 
As the combined multi-source domain contains more information than each single domain, 
it can reinforce the nonspontaneous common information shared across multiple source domains. 
We hence use the $(N+1)$-th subnetwork as a guidance network and use
the first $N$ subnetworks as branch networks 
in our proposed mutual learning framework.

	For each branch network, the corresponding source domain data and the target domain data are used as inputs.
	The combined multiple source domain data and the target domain data are used as inputs for the guidance network.
	All these subnetworks have the same structure that has three components: 
	feature extractor $G$, domain discriminator $D$, and category classifier $F$.  
	The parameters of the first few layers in the feature extractors are shared across all the subnetworks
	to enable common low-level feature extraction, while the remaining layers are separated to 
	capture source-domain specific information. 
	For each subnetwork, the input data first go through the feature extraction network to produce high level features.
	Source domain dependent conditional adversarial feature alignment is then conducted to align feature distributions 
	between each specific source domain (or combined source domains) and target domain
	using a separate domain discriminator as an adversary under an adversarial loss $L_{adv}$.
	The classifiers predict the class labels of the input samples based on the aligned features 
	with classification losses $L_C$ and $L_E$,
	while mutual learning
	is conducted by enforcing prediction distribution alignment between each branch network 
	and the guidance network on corresponding samples.
	A prediction misalignment loss $L_M$ is considered between each branch network and the guidance network.
	Below we present these loss terms. 

\subsection{Conditional Adversarial Feature Alignment}. 
We propose to deploy conditional adversarial domain adaptation to align 
feature distributions between the source domain and the target domain 
and induce domain invariant features.
As stated above, all the $N+1$ adaptation subnetworks share the same structure. 
Hence the conditional adversarial feature alignment is conducted in the same manner for different subnetworks.
The fundamental difference is that different subnetworks use different source domain data as input
and the adversarial alignment results will be source domain dependent.
Here we take the $j$-th subnetwork as an example to present the conditional adversarial feature alignment
adopted in the proposed model.

The main idea of adversarial domain adaptation is to 
adopt the adversarial learning principle of generative adversarial networks
into the domain adaptation setting by introducing an adversary domain discriminator $D$~\cite{ganin2016domain}.
For the $j$-th subnetwork, this implies playing a minimax game between the feature extractor $G_j$ 
and the domain discriminator $D_j$,
where $D_j$ tries to maximumly distinguish the source domain data $G_j({\bf X}_{S_j})$ 
from the target domain data $G_j({\bf X}_{T})$ 
and $G_j$ tries to maximumly deceive the discriminator. 
Moreover, although we like to drop the domain divergence, it is important to improve the discriminability of the induced features
towards the final classification task. 
We hence take the classifier's label prediction results into account to perform conditional adversarial 
domain adaptation with the following adversarial loss:
\begin{align}
	{L_{adv_j}} =& 
	\frac{1}{n_s^j}\sum\limits_{i=1}^{n_s^j} 
	\log\! \left[{{D_j}\!\left( \Phi(G_j({\bf x}_i^j), {\bf p}_i^j) \right)\!} \right] + \nonumber\\
	&\frac{1}{n_t}\sum\limits_{i=1}^{n_t} 
	\log\! \left[1-{{D_j}\!\left( \Phi(G_j({\bf x}_i^t), {\bf p}_i^{t_j}) \right)\!} \right] 
\end{align}
where ${\bf p}_i^j$ is the prediction probability vector produced by the classifier $F_j$ on image ${\bf x}_i^j$,
such that 
\begin{align}
{\bf p}_i^j=F_j(G_j({\bf x}_i^j)),\qquad
	{\bf p}_i^{t_j}=F_j(G_j({\bf x}_i^t)),
\end{align}
For $K$-class classification problem, ${\bf p}_i^j$ will be a length $K$ vector with each entry indicating
the probability of  ${\bf x}_i^j$ belonging to the corresponding class category.
$\Phi(\cdot,\cdot)$ denotes the conditioning strategy function. For simplicity, one can use a simple concatenation
$\Phi({\bf q},{\bf p})=[{\bf q};{\bf p}]$. 
In this work, we used the multilinear conditioning function proposed in~\cite{long2018conditional},
as it can capture the cross covariance between feature representations and classifier predictions
to help preserve the discriminability of the features.

Finally the overall adversarial loss from all the $N+1$ subnetworks can be computed as:
\begin{align}
	{L_{adv}} = \frac{1}{N+1}\sum_{j=1}^{N+1}{L_{adv_j}} 
\end{align}

\subsection{Semi-Supervised Prediction Loss}
Following the structure of ML-MSDA in Figure~\ref{fig:ML-MSDA},
the extracted domain invariant features in each subnetwork will be served as input to the classifier $F_j$.
For the labeled images from the source domain, we can use the supervised cross-entropy loss 
to perform training: 

\begin{align}
	L_C=-\frac{1}{N+1}\sum_{j=1}^{N+1}\Big(\frac{1}{n_s^j}\sum_{i=1}^{n_s^j}{\bf y}_i^{j\top}\log 
	{\bf p}_i^j\Big) 
\end{align}
For the unlabeled data from the target domain, we use an unsupervised entropy loss 
to include them into the classifier training:
\begin{align}
	L_E=-\frac{1}{N+1}\sum_{j=1}^{N+1}\Big(\frac{1}{n_t}\sum_{i=1}^{n_t}{\bf p}_i^{t_j\top}\log 
	{\bf p}_i^{t_j}\Big)
\end{align}
The assumption is that if the source and target domains are well aligned, the classifier trained on the labeled source images
should be able to make confident predictions on the target domain images and hence have small prediction entropy values.
Therefore we expect this entropy loss can help bridge domain divergence and induce discriminative features. 

\subsection{Guidance Network Centered Mutual Learning}. 
With the adversarial feature alignment 
in each branch network, the target domain is aligned with each source domain separately. 
Due to the existence of domain shift among the multiple source domains, 
the domain invariant features extracted and the consequent classifier trained in one subnetwork 
will be different from that in another subnetwork.  
Nevertheless, under effective domain adaptation, 
the divergence between each subnetwork's prediction result on the target domain data
and the true labels should be small.
By sharing the same target domain, this implies the prediction results of all the subnetworks
in the target domain should be consistent. 
%
Under this assumption,
in order to improve the generalization performance of the model 
and increase the robustness of network training,
we propose to conduct mutual learning over all the subnetworks by minimizing their prediction inconsistency 
in the shared target domain. 
As the guidance network used data from all the source domains as a combined domain, 
it contains more transferable information than each branch network.
Hence we propose to 
enforce prediction consistency 
by aligning each branch network
with the guidance network in terms of predicted label distribution for each target instance.
Specifically, we can use the Kullback Leibler (KL) Divergence to 
align the predicted label probability vector for each target domain instance 
from the {\em j}-th branch network with the predicted label probability vector for the same instance 
from the guidance network; that is,
\begin{align}
	\mathcal{D}_{KL}({\bf p}_i^{t_j}\|{\bf p}_i^{t_{N+1}}) ={\bf p}_i^{t_j\top}[\log{\bf p}_i^{t_j} -\log{\bf p}_i^{t_{N+1}} ]
\end{align}
where \({{\textbf{p}}_i^{{t_j}}}\) 
and \({{\textbf{p}}_i^{{t_{N+1}}}}\) 
are the predicted label probability vectors for the $i$-th instance in the target domain
produced by the $j$-th branch network and the guidance network respectively.
Since the KL divergence metric is asymmetric, we use a symmetric Jensen-Shannon Divergence loss \cite{ZhangDeep} instead,
which leads to the following overall prediction inconsistency loss:
\begin{align}
	\!\!\!	L_M  =\frac{1}{2Nn_t}\sum_{j=1}^N\sum_{i=1}^{n_t}
	\left[
	\mathcal{D}_{KL}({\bf p}_i^{t_j}\|{\bf p}_i^{t_{N+1}}) + 
	\mathcal{D}_{KL}({\bf p}_i^{t_{N+1}}\|{\bf p}_i^{t_{j}}) 
	\right]
\end{align}
This loss enforces regularizations over the prediction inconsistency on the target domain instances 
across the multiple subnetworks.

\subsection{Overall Learning Problem and Prediction}
By integrating the prediction loss, adversarial feature alignment loss, and the 
prediction inconsistency loss together, 
we have the following overall adversarial learning problem:
\begin{align}
\mathop {\min }\limits_{G,F} \mathop {\max }\limits_D \quad {L_C} + \alpha {L_M} + \beta {L_E} + \lambda {L_{adv}}
\end{align}
where \(\alpha \), \(\beta \) and  \(\lambda \) are trade-off hyperparameters;
$G, F$ and $D$ denote the sets of $N+1$ feature extractors, classifiers and domain discriminators respectively.
This training problem can be solved using standard stochastic gradient descent algorithms
by performing min-max adversarial updates. 

After training, we obtain 
{\em N}+1 classifiers from the model.
We use these classifiers to predict the labels of the unlabeled target domain instances
in a guidance network centered ensemble manner. 
For the $i$-th instance in the target domain, 
the ensemble prediction probability result is:
\begin{align}
	{\bf p}_i^{t} = \frac{1}{2}\Big({\bf p}_i^{t_{N+1}} +\frac{1}{N}\sum_{j=1}^N{\bf p}_i^{t_j}
	\Big)
\end{align}
where the prediction from the guidance network is qiven equal weight to the average prediction results
from the other $N$ branch networks.
%


\begin{table*}[th]
\begin{center}
\vskip .1in	
\setlength{\abovecaptionskip}{-10pt}
\setlength{\belowcaptionskip}{-10pt}
\caption{{Test results on Digit Recognition.} The average classification accuracy of the proposed approach is 90.68\%, which is 3.03\% higher than the best comparison method.}
\vskip .1in	
\setlength{\tabcolsep}{5pt}
\resizebox{0.9\textwidth}{!}{
\begin{tabular}{c|c|c|c|c|c|c|c}
\toprule
Standards & Models & \tabincell{c}{\textbf{mt,up,sv,}\\\textbf{sy}\(\to\)\textbf{mm}} & \tabincell{c}{\textbf{mm,up,sv,}\\\textbf{sy}\(\to\)\textbf{mt}} & \tabincell{c}{\textbf{mm,mt,sv,}\\\textbf{sy}\(\to\)\textbf{up}} & \tabincell{c}{\textbf{mm,mt,up,}\\\textbf{sy}\(\to\)\textbf{sv}} & \tabincell{c}{\textbf{mm,mt,up,}\\\textbf{sv}\(\to\)\textbf{sy}} & Avg \\ \hline
\multirowcell{3}{Source\\Combine} & Source Only & 63.70\scriptsize{\(\pm\)0.83} & 92.30\scriptsize{\(\pm\)0.91} & 90.71\scriptsize{\(\pm\)0.54} & 71.51\scriptsize{\(\pm\)0.75} & 83.44\scriptsize{\(\pm\)0.79} & 80.33\scriptsize{\(\pm\)0.76} \\
 & DAN~\cite{long2015learning} & 67.87\scriptsize{\(\pm\)0.75} & 97.50\scriptsize{\(\pm\)0.62} & 93.49\scriptsize{\(\pm\)0.85} & 67.80\scriptsize{\(\pm\)0.84} & 86.93\scriptsize{\(\pm\)0.93} & 82.72\scriptsize{\(\pm\)0.79} \\
 & DANN~\cite{ganin2014unsupervised} & 70.81\scriptsize{\(\pm\)0.94} & 97.90\scriptsize{\(\pm\)0.83} & 93.47\scriptsize{\(\pm\)0.79} & 68.50\scriptsize{\(\pm\)0.85} & 87.37\scriptsize{\(\pm\)0.68} & 83.61\scriptsize{\(\pm\)0.82} \\ \hline
\multirowcell{12}{Multi-\\Source} & Source Only & 63.37\scriptsize{\(\pm\)0.74} & 90.50\scriptsize{\(\pm\)0.83} & 88.71\scriptsize{\(\pm\)0.89} & 63.54\scriptsize{\(\pm\)0.93} & 82.44\scriptsize{\(\pm\)0.65} & 77.71\scriptsize{\(\pm\)0.81} \\
 & DAN~\cite{long2015learning} & 63.78\scriptsize{\(\pm\)0.71} & 96.31\scriptsize{\(\pm\)0.54} & 94.24\scriptsize{\(\pm\)0.87} & 62.45\scriptsize{\(\pm\)0.72} & 85.43\scriptsize{\(\pm\)0.77} & 80.44\scriptsize{\(\pm\)0.72} \\
 & CORAL~\cite{sun2016return} & 62.53\scriptsize{\(\pm\)0.69} & 97.21\scriptsize{\(\pm\)0.83} & 93.45\scriptsize{\(\pm\)0.82} & 64.40\scriptsize{\(\pm\)0.72} & 82.77\scriptsize{\(\pm\)0.69} & 80.07\scriptsize{\(\pm\)0.75} \\
 & DANN~\cite{ganin2014unsupervised} & 71.30\scriptsize{\(\pm\)0.56} & 97.60\scriptsize{\(\pm\)0.75} & 92.33\scriptsize{\(\pm\)0.85} & 63.48\scriptsize{\(\pm\)0.79} & 85.34\scriptsize{\(\pm\)0.84} & 82.01\scriptsize{\(\pm\)0.76} \\
 & JAN~\cite{long2017deep} & 65.88\scriptsize{\(\pm\)0.68} & 97.21\scriptsize{\(\pm\)0.73} & 95.42\scriptsize{\(\pm\)0.77} & 75.27\scriptsize{\(\pm\)0.71} & 86.55\scriptsize{\(\pm\)0.64} & 84.07\scriptsize{\(\pm\)0.71} \\
 & ADDA~\cite{tzeng2017adversarial} & 71.57\scriptsize{\(\pm\)0.52} & 97.89\scriptsize{\(\pm\)0.84} & 92.83\scriptsize{\(\pm\)0.74} & 75.48\scriptsize{\(\pm\)0.48} & 86.45\scriptsize{\(\pm\)0.62} & 84.84\scriptsize{\(\pm\)0.64} \\
 & DCTN~\cite{xu2018deep} & 70.53\scriptsize{\(\pm\)1.24} & 96.23\scriptsize{\(\pm\)0.82} & 92.81\scriptsize{\(\pm\)0.27} & 77.61\scriptsize{\(\pm\)0.41} & 86.77\scriptsize{\(\pm\)0.48} & 84.79\scriptsize{\(\pm\)0.27} \\
 & MEDA~\cite{wang2018visual} & 71.31\scriptsize{\(\pm\)0.75} & 96.47\scriptsize{\(\pm\)0.78} & 97.01\scriptsize{\(\pm\)0.82} & 78.45\scriptsize{\(\pm\)0.77} & 84.62\scriptsize{\(\pm\)0.79} & 85.60\scriptsize{\(\pm\)0.78} \\
 & MCD~\cite{saito2018maximum} & 72.50\scriptsize{\(\pm\)0.67} & 96.21\scriptsize{\(\pm\)0.81} & 95.33\scriptsize{\(\pm\)0.74} & 78.89\scriptsize{\(\pm\)0.78} & 87.47\scriptsize{\(\pm\)0.65} & 86.10\scriptsize{\(\pm\)0.73} \\
  & \({{{\rm{M}}^{\rm{3}}}{\rm{SDA}}}\)~\cite{peng2019moment} & 69.76\scriptsize{\(\pm\)0.86} & 98.58\scriptsize{\(\pm\)0.47} & 95.23\scriptsize{\(\pm\)0.79} & 78.56\scriptsize{\(\pm\)0.95} & 87.56\scriptsize{\(\pm\)0.53} & 86.13\scriptsize{\(\pm\)0.64} \\
	& \({{{\rm{M}}^{\rm{3}}}{\rm{SDA\!-\!\beta}}}\)~\cite{peng2019moment} & 72.82\scriptsize{\(\pm\)1.13} & 98.43\scriptsize{\(\pm\)0.68} & 96.14\scriptsize{\(\pm\)0.81} & \textbf{81.32}\scriptsize{\(\pm\)0.86} & {\bf 89.58}\scriptsize{\(\pm\)0.56} & 87.65\scriptsize{\(\pm\)0.75} \\
 & \tabincell{c}{ML-MSDA\\(ours)}& \textbf{96.62}\scriptsize{\(\pm\)0.15} & \textbf{99.37}\scriptsize{\(\pm\)0.06} & \textbf{98.29}\scriptsize{\(\pm\)0.13} & 70.27\scriptsize{\(\pm\)0.64} & {88.52}\scriptsize{\(\pm\)1.29} & \textbf{90.68}\scriptsize{\(\pm\)0.46} \\ \bottomrule
\end{tabular}}
\label{table:digit-five}
\end{center}
\end{table*}
%
\section{Experiments}
To investigate the effectiveness of the proposed approach, 
we conducted experiments on three well-known benchmark multi-source domain adaptation datasets:
Digit-five dataset, OfficeCaltech10 dataset and DomainNet dataset. 
We compared the proposed ML-MSDA with the state-of-the-art UDA and MSDA methods,
and report the comparison results in this section.
\\

\noindent{\bf Implementation Details.}
The experiments are conducted using PyTorch. 
For the proposed ML-MSDA we set the trade-off hyperparameters (\(\lambda \), \(\alpha \), \(\beta \)) 
as (5, 5, 0.5) respectively. 
We define the process of training on all samples of the combined-source domain as an epoch. 
The learning rate is set as 0.01 for the first 10 epochs and 
as 0.001 in the following 10 epochs. 
After the first 20 epochs, the learning rate is set as 0.0001. 
In the experiments on Digit Recognition, each batch is composed of 256 samples. 
On Office-Caltech10 and DomainNet, we set the batch-size as 20 due to the large size of images.

\subsection{Experiments on Digit Recognition}
The Digit Recognition dataset consists of 10  classes of digit images sampled from five different datasets, 
including \textbf{mt} ({\em MNIST})~\cite{lecun1998gradient}, \textbf{mm} ({\em MNIST-M})~\cite{lecun1998gradient}, \textbf{sv} ({\em SVHN}), \textbf{up} ({\em USPS}), and \textbf{sy} ({\em Synthetic Digits})~\cite{ganin2014unsupervised}, 
which form five domains.  
Following previous studies, \(\it{M^3}\it{SDA}\)~\cite{peng2019moment} and \textit{DCTN}~\cite{xu2018deep},
 on multi-source domain adaptation, 
we randomly chose 25,000 images for training and 9000 for testing in {\em MNIST}, {\em MNIST-M}, and {\em SVHN}. 
For small datasets {\em USPS} and {\em Synthetic Digits}, 
we used all their training and testing samples. 
With these five datasets,
five domain adaptation tasks are naturally formed by selecting one dataset as the target domain
and using the others as the source domains in turn.

We compared the proposed ML-MSDA method with two state-of-the-art MSDA approaches,
Moment Matching for Multi-Source Domain Adaptation (\(\it{M^3}\it{SDA}\))~\cite{peng2019moment} 
and Deep Cocktail Network (\textit{DCTN})~\cite{xu2018deep}.
In addition, we also compared with a number of UDA methods 
including
Deep Alignment Network (\textit{DAN})~\cite{long2015learning}, 
Domain Adversarial Neural Network (\textit{DANN})~\cite{ganin2014unsupervised}, 
Correlation Alignment (\textit{CORAL})~\cite{sun2016return}, 
Joint Adaptation Network (\textit{JAN})~\cite{long2017deep}, 
Adversarial Discriminative Domain Adaptation (\textit{ADDA})~\cite{tzeng2017adversarial}, 
Manifold Embedded Distribution Alignment (\textit{MEDA})~\cite{wang2018visual}, 
and Maximum Classifier Discrepancy (MCD)~\cite{saito2018maximum}. 
Following experiments in previous multi-source DA study,
for these single-source UDA methods, 
we recorded the averages of their multiple single-source domain adaptation results
under the multi-source setting.

Following the backbone network setting of~\cite{peng2019moment},
we used three {\em conv} layers and two {\em fc} layers as the feature extractor 
and a single {\em fc} layer as the category classifier.  
As the model is small, we did not use weight sharing across different branches.
The same backbone network was used in all the experiments.
We repeat each experiment five times, and report
the mean and standard deviation values of the test accuracy results in the target domain. 

The comparison results are reported in Table \ref{table:digit-five}. 
We can see ML-MSDA outperforms all the other methods on three out of the five 
domain adaptation tasks.
The average test accuracy of the proposed ML-MSDA method across the five domain adaptation
tasks is 90.68\%,
which outperforms 
the best alterative multi-source domain adaptation method, \(M^3\)SDA-$\beta$, and 
all the other comparison methods with notable performance gains. 
These results suggest the proposed mutual learning network model is very effective.

\subsection{Experiments on Office-Caltech10}
Office-Caltech10 dataset~\cite{gong2012geodesic} is collected from four different domains: \textbf{A} ({\em Amazon}), \textbf{C} ({\em Caltech}), \textbf{W} ({\em Webcam}) and \textbf{D} ({\em DSLR}). It consists of 10 object categories, 
and each domain includes 958, 295, 157, and 1,123 images, respectively. 
On this dataset, four domain adaptation tasks are constructed by using one domain as the target domain in turn 
and the others as source domains.

We compared the results produced by the proposed ML-MSDA method
with the results of a number of state-of-the-art domain adaptation methods, including
\textit{DAN}~\cite{long2015learning}, \textit{DCTN}~\cite{xu2018deep}, \textit{JAN}~\cite{long2017deep}, \textit{MEDA}~\cite{wang2018visual}, \textit{MCD}~\cite{saito2018maximum} and \(\textit{M}^\textit{3}\textit{SDA}\)\cite{peng2019moment}.
For fair comparison, we used ResNet101 pre-trained on ImageNet as the backbone network in all the experiments. 
For ML-MSDA, the weights of {\em conv}1, {\em conv}2 and {\em conv}3 stages are shared among the guidance network and all branch networks. 
But each network trains their {\em conv}4 and {\em conv}5 stages separately. 

The comparison results on Office-Caltech10 are reported in Table \ref{table:office-10}. 
We can see that on this small dataset, all domain adaptation methods work very well.
Nevertheless, our proposed ML-MSDA consistently outperforms all other methods
and achieves a 97.6\% average accuracy.

\begin{table}[t!]
\vskip .1in	
\begin{center}
\setlength{\abovecaptionskip}{-10pt}
\setlength{\belowcaptionskip}{-10pt}
\caption{{Results on Office-Caltech10.} 
The average classification accuracy of the proposed approach is 97.6\%, which is 1.2\% higher than the best comparison result.}
\begin{small}
\setlength{\tabcolsep}{1pt}
\vskip .1in	
\resizebox{.48\textwidth}{!}{
\begin{tabular}{c|c|c|c|c|c|c}
\toprule
Standards & Models & \tabincell{c}{\textbf{A,C,D}\\\(\to\)\textbf{W}} & \tabincell{c}{\textbf{A,C,W}\\\(\to\)\textbf{D}} & \tabincell{c}{\textbf{A,D,W}\\\(\to\)\textbf{C}} & \tabincell{c}{\textbf{C,D,W}\\\(\to\)\textbf{A}} & Avg \\ \hline
\multirowcell{2}{Source\\Combine} & Source only & 99 & 98.3 & 87.8 & 86.1 & 92.8 \\
 & DAN~\cite{long2015learning} & 99.3 & 98.2 & 89.7 & 94.8 & 95.5 \\ \hline
\multirowcell{9}{Multi-\\Source} & Source only & 99.1 & 98.2 & 85.4 & 88.7 & 92.9 \\
 & DAN~\cite{long2015learning} & 99.5 & 99.1 & 89.2 & 91.6 & 94.8 \\
 & DCTN~\cite{xu2018deep} & 99.4 & 99 & 90.02 & 92.7 & 95.3 \\
 & JAN~\cite{long2017deep} & 99.4 & 99.4 & 91.2 & 91.8 & 95.5 \\
 & MEDA~\cite{wang2018visual} & 99.3 & 99.2 & 91.4 & 92.9 & 95.7 \\
 & MCD~\cite{saito2018maximum} & 99.5 & 99.1 & 91.5 & 92.1 & 95.6 \\
 & \({{{\rm{M}}^{\rm{3}}}{\rm{SDA}}}\)~\cite{peng2019moment} & 99.4 & 99.2 & 91.5 & 94.1 & 96.1 \\
 & \({{{\rm{M}}^{\rm{3}}}{\rm{SDA-\beta}}}\)~\cite{peng2019moment} & 99.5 & 99.2 & 92.2 & 94.5 & 96.4 \\
 & ML-MSDA (ours) & \textbf{100} & \textbf{100} & \textbf{94.7} & \textbf{95.7} & \textbf{97.6} \\ \hline
\end{tabular}}
\end{small}
\label{table:office-10}
\end{center}
\end{table}

\begin{table}[t]
\setlength{\abovecaptionskip}{-10pt}
\setlength{\belowcaptionskip}{-10pt}
\caption{{Details of DomainNet dataset.} The ratio of train/test is 70\%/30\%.}
\begin{center}
\renewcommand\arraystretch{1.10}
\setlength{\tabcolsep}{1pt}
\resizebox{.48\textwidth}{!}{
\begin{tabular}{c|c|c|c|c|c|c|c}
\toprule
 & \textbf{clp} & \textbf{inf} & \textbf{pnt} & \textbf{qdr} & \textbf{rel} & \textbf{skt} & \textbf{Total} \\ \hline
Train & 34,019 & 37,087 & 52,867 & 120,750 & 122,563 & 49,115 & 416,401 \\
Test & 14,818 & 16,114 & 22,892 & 51,750 & 52,764 & 21,271 & 179,609 \\ \hline
Total & 48,837 & 53,201 & 75,759 & 172,500 & 175,327 & 70,386 & 596,010 \\
Per-Class & 141 & 154 & 219 & 500 & 508 & 204 & 1,728 \\ \bottomrule
\end{tabular}}
\end{center}
\label{table:DomainNet}
\end{table}
\begin{table*}[t]
\vskip .1in	
\begin{center}
\setlength{\abovecaptionskip}{-10pt}
\setlength{\belowcaptionskip}{-10pt}
\caption{{Results on DomainNet dataset.} The proposed ML-MSDA produced the best 
	average accuracy 44.3\% among the domain adaptation methods.}
\vskip .1in	
\setlength{\tabcolsep}{5pt}
\resizebox{.92\textwidth}{!}{
\begin{tabular}{c|c|c|c|c|c|c|c|c}
\toprule
Standards & Models & \tabincell{c}{\textbf{inf,pnt,}\\\textbf{qdr,rel,}\\\textbf{skt}\(\to\)\textbf{clp}} & \tabincell{c}{\textbf{clp,pnt,}\\\textbf{qdr,rel,}\\\textbf{skt}\(\to\)\textbf{inf}} & \tabincell{c}{\textbf{clp,inf,}\\\textbf{qdr,rel,}\\\textbf{skt}\(\to\)\textbf{pnt}} & \tabincell{c}{\textbf{clp,inf,}\\\textbf{pnt,rel,}\\\textbf{skt}\(\to\)\textbf{qdr}} & \tabincell{c}{\textbf{clp,inf,}\\\textbf{pnt,qdr,}\\\textbf{skt}\(\to\)\textbf{rel}} & \tabincell{c}{\textbf{clp,inf,}\\\textbf{pnt,qdr,}\\\textbf{rel}\(\to\)\textbf{skt}} & Avg \\ \hline
 \multirowcell{8}{Single\\Best} & Source Only & 39.6\scriptsize{\(\pm\)0.58} & 8.2\scriptsize{\(\pm\)0.75} & 33.9\scriptsize{\(\pm\)0.62} & 11.8\scriptsize{\(\pm\)0.69} & 41.6\scriptsize{\(\pm\)0.84} & 23.1\scriptsize{\(\pm\)0.72} & 26.4\scriptsize{\(\pm\)0.70} \\
 & DAN~\cite{long2015learning} & 39.1\scriptsize{\(\pm\)0.51} & 11.4\scriptsize{\(\pm\)0.81} & 33.3\scriptsize{\(\pm\)0.62} & 16.2\scriptsize{\(\pm\)0.38} & 42.1\scriptsize{\(\pm\)0.73} & 29.7\scriptsize{\(\pm\)0.93} & 28.6\scriptsize{\(\pm\)0.63} \\
 & RTN~\cite{long2016unsupervised} & 35.3\scriptsize{\(\pm\)0.73} & 10.7\scriptsize{\(\pm\)0.61} & 31.7\scriptsize{\(\pm\)0.82} & 13.1\scriptsize{\(\pm\)0.68} & 40.6\scriptsize{\(\pm\)0.55} & 26.6\scriptsize{\(\pm\)0.78} & 26.3\scriptsize{\(\pm\)0.70} \\
 & JAN~\cite{long2017deep} & 35.3\scriptsize{\(\pm\)0.71} & 9.1\scriptsize{\(\pm\)0.63} & 32.5\scriptsize{\(\pm\)0.65} & 14.3\scriptsize{\(\pm\)0.62} & 43.1\scriptsize{\(\pm\)0.78} & 25.7\scriptsize{\(\pm\)0.61} & 26.7\scriptsize{\(\pm\)0.67} \\
 & DANN~\cite{ganin2014unsupervised} & 37.9\scriptsize{\(\pm\)0.69} & 11.4\scriptsize{\(\pm\)0.91} & 33.9\scriptsize{\(\pm\)0.60} & 13.7\scriptsize{\(\pm\)0.56} & 41.5\scriptsize{\(\pm\)0.67} & 28.6\scriptsize{\(\pm\)0.63} & 27.8\scriptsize{\(\pm\)0.68} \\
 & ADDA~\cite{tzeng2017adversarial} & 39.5\scriptsize{\(\pm\)0.81} & 14.5\scriptsize{\(\pm\)0.69} & 29.1\scriptsize{\(\pm\)0.78} & 14.9\scriptsize{\(\pm\)0.54} & 41.9\scriptsize{\(\pm\)0.82} & 30.7\scriptsize{\(\pm\)0.68} & 28.4\scriptsize{\(\pm\)0.72} \\
 & SE~\cite{french2017self} & 31.7\scriptsize{\(\pm\)0.70} & 12.9\scriptsize{\(\pm\)0.58} & 19.9\scriptsize{\(\pm\)0.75} & 7.7\scriptsize{\(\pm\)0.44} & 33.4\scriptsize{\(\pm\)0.56} & 26.3\scriptsize{\(\pm\)0.50} & 22.0\scriptsize{\(\pm\)0.66} \\
 & MCD~\cite{saito2018maximum} & 42.6\scriptsize{\(\pm\)0.32} & 19.6\scriptsize{\(\pm\)0.76} & 42.6\scriptsize{\(\pm\)0.98} & 3.8\scriptsize{\(\pm\)0.64} & 50.5\scriptsize{\(\pm\)0.43} & 33.8\scriptsize{\(\pm\)0.89} & 32.2\scriptsize{\(\pm\)0.66} \\ \hline
\multirowcell{8}{Source\\Combine} & Source only & 47.6\scriptsize{\(\pm\)0.52} & 13.0\scriptsize{\(\pm\)0.41} & 38.1\scriptsize{\(\pm\)0.45} & 13.3\scriptsize{\(\pm\)0.39} & 51.9\scriptsize{\(\pm\)0.85} & 33.7\scriptsize{\(\pm\)0.54} & 32.9\scriptsize{\(\pm\)0.54} \\
 & DAN~\cite{long2015learning} & 45.4\scriptsize{\(\pm\)0.49} & 12.8\scriptsize{\(\pm\)0.86} & 36.2\scriptsize{\(\pm\)0.58} & 15.3\scriptsize{\(\pm\)0.37} & 48.6\scriptsize{\(\pm\)0.72} & 34.0\scriptsize{\(\pm\)0.54} & 32.1\scriptsize{\(\pm\)0.59} \\
 & RTN~\cite{long2016unsupervised} & 44.2\scriptsize{\(\pm\)0.57} & 12.6\scriptsize{\(\pm\)0.73} & 35.3\scriptsize{\(\pm\)0.59} & 14.6\scriptsize{\(\pm\)0.76} & 48.4\scriptsize{\(\pm\)0.67} & 31.7\scriptsize{\(\pm\)0.73} & 31.1\scriptsize{\(\pm\)0.68} \\
 & JAN~\cite{long2017deep} & 40.9\scriptsize{\(\pm\)0.43} & 11.1\scriptsize{\(\pm\)0.61} & 35.4\scriptsize{\(\pm\)0.50} & 12.1\scriptsize{\(\pm\)0.67} & 45.8\scriptsize{\(\pm\)0.59} & 32.3\scriptsize{\(\pm\)0.63} & 29.6\scriptsize{\(\pm\)0.57} \\
 & DANN~\cite{ganin2014unsupervised} & 45.5\scriptsize{\(\pm\)0.59} & 13.1\scriptsize{\(\pm\)0.72} & 37.0\scriptsize{\(\pm\)0.69} & 13.2\scriptsize{\(\pm\)0.77} & 48.9\scriptsize{\(\pm\)0.65} & 31.8\scriptsize{\(\pm\)0.62} & 32.6\scriptsize{\(\pm\)0.68} \\
 & ADDA~\cite{tzeng2017adversarial} & 47.5\scriptsize{\(\pm\)0.76} & 11.4\scriptsize{\(\pm\)0.67} & 36.7\scriptsize{\(\pm\)0.53} & 14.7\scriptsize{\(\pm\)0.50} & 49.1\scriptsize{\(\pm\)0.82} & 33.5\scriptsize{\(\pm\)0.49} & 32.2\scriptsize{\(\pm\)0.63} \\
 & SE~\cite{french2017self} & 24.7\scriptsize{\(\pm\)0.32} & 3.9\scriptsize{\(\pm\)0.47} & 12.7\scriptsize{\(\pm\)0.35} & 7.1\scriptsize{\(\pm\)0.46} & 22.8\scriptsize{\(\pm\)0.51} & 9.1\scriptsize{\(\pm\)0.49} & 16.1\scriptsize{\(\pm\)0.43} \\
 & MCD~\cite{saito2018maximum} & 54.3\scriptsize{\(\pm\)0.64} & 22.1\scriptsize{\(\pm\)0.70} & 45.7\scriptsize{\(\pm\)0.63} & 7.6\scriptsize{\(\pm\)0.49} & 58.4\scriptsize{\(\pm\)0.65} & 43.5\scriptsize{\(\pm\)0.57} & 38.5\scriptsize{\(\pm\)0.61} \\ \hline
\multirowcell{4}{Multi-\\Source} & DCTN~\cite{xu2018deep} & 48.6\scriptsize{\(\pm\)0.73} & 23.5\scriptsize{\(\pm\)0.59} & 48.8\scriptsize{\(\pm\)0.63} & 7.2\scriptsize{\(\pm\)0.46} & 53.5\scriptsize{\(\pm\)0.56} & 47.3\scriptsize{\(\pm\)0.47} & 38.2\scriptsize{\(\pm\)0.57} \\
& \({{{\rm{M}}^{\rm{3}}}{\rm{SDA}}}\)~\cite{peng2019moment} & 57.2\scriptsize{\(\pm\)0.98} & 24.2\scriptsize{\(\pm\)1.21} & 51.6\scriptsize{\(\pm\)0.44} & 5.2\scriptsize{\(\pm\)0.45} & 61.6\scriptsize{\(\pm\)0.89} & 49.6\scriptsize{\(\pm\)0.56} & 41.5\scriptsize{\(\pm\)0.74} \\
 & \({{{\rm{M}}^{\rm{3}}}{\rm{SDA\!-\!\beta}}}\)\cite{peng2019moment} & 58.6\scriptsize{\(\pm\)0.53} & 26.0\scriptsize{\(\pm\)0.89} & \textbf{52.3}\scriptsize{\(\pm\)0.55} & 6.3\scriptsize{\(\pm\)0.58} & \textbf{62.7}\scriptsize{\(\pm\)0.51} & 49.5\scriptsize{\(\pm\)0.76} & 42.6\scriptsize{\(\pm\)0.64} \\
 & \tabincell{c}{ML-MSDA\\(ours)} & \textbf{61.4}\scriptsize{\(\pm\)0.79} & \textbf{26.2}\scriptsize{\(\pm\)0.41} & 51.9\scriptsize{\(\pm\)0.20} & \textbf{19.1}\scriptsize{\(\pm\)0.31} & 57.0\scriptsize{\(\pm\)1.04} & \textbf{50.3}\scriptsize{\(\pm\)0.67} & \textbf{44.3}\scriptsize{\(\pm\)0.57} \\ \hline
\multirowcell{3}{Oracle\\Results} & AlexNet & 65.5\scriptsize{\(\pm\) 0.56} & 27.7\scriptsize{\(\pm\)0.34} & 57.6\scriptsize{\(\pm\)0.49} & 68.0\scriptsize{\(\pm\)0.55} & 72.8\scriptsize{\(\pm\)0.67} & 56.3\scriptsize{\(\pm\)0.59} & 58.0\scriptsize{\(\pm\)0.53} \\
 & ResNet101 & 69.3\scriptsize{\(\pm\)0.37} & 34.5\scriptsize{\(\pm\)0.42} & 66.3\scriptsize{\(\pm\)0.67} & 66.8\scriptsize{\(\pm\)0.51} & 80.0\scriptsize{\(\pm\)0.59} & 60.7\scriptsize{\(\pm\)0.48} & 63.0\scriptsize{\(\pm\)0.51} \\
 & ResNet152 & 71.0\scriptsize{\(\pm\)0.63} & 36.1\scriptsize{\(\pm\)0.61} & 68.1\scriptsize{\(\pm\)0.49} & 69.1\scriptsize{\(\pm\)0.52} & 81.3\scriptsize{\(\pm\)0.49} & 65.2\scriptsize{\(\pm\)0.57} & 65.1\scriptsize{\(\pm\)0.55} \\ 
 \bottomrule
\end{tabular}}
\label{table:result-domainnet}
\end{center}
\vskip -.1in
\end{table*}

\begin{table*}[]
\begin{center}
\vskip .1in	
\caption{{Ablation study.} 
	Comparison of the proposed approach with its five variants. }
\vskip .1in	
\setlength{\tabcolsep}{8pt}
\resizebox{.75\textwidth}{!}{
\begin{tabular}{l|c|c|c|c}
\toprule
 & \tabincell{c}{{\textbf{mt, up, sv,}}\\ {\textbf{sy}\(\to\)\textbf{mm}}}& \tabincell{c}{{\textbf{D,W,C}}\\ {\(\to\)\textbf{A}}} & \tabincell{c}{{\textbf{clp, inf, qdr,}}\\ {\textbf{rel, skt}\(\to\)\textbf{pnt}}} &  \tabincell{c}{{\textbf{inf, pnt, qdr,}}\\ {\textbf{rel, skt}\(\to\)\textbf{clp}}} \\ \hline
 \tabincell{c}{ML-w/o condition-adv}& 92.1 & 95.5 & 50.5 & 58.3 \\ \hline
  \tabincell{c}{ML-w/o \(L_{E}\)}& 94.5 & 95.4 & 44.3 & 51.1 \\ \hline
 \tabincell{c}{ML-w/o \(L_{M}\)}& 91.9 & 94.1 & 43.3 & 56.9\\ \hline
 \tabincell{c}{ML-guidance-inf} & 95.7 & 95.8 & 51.0 & 58.7\\ \hline
 ML-branch-average-inf& 96.0 & 95.4 & 41.3 & 48.4\\ \hline
	ML-MSDA (full) & 96.6 & 95.7 & 51.9 & 61.0\\
 \bottomrule
\end{tabular}
}
\label{table:ablation}
\end{center}
\end{table*}

\begin{figure*}[t]
\begin{center}
\includegraphics[width=0.90\linewidth]{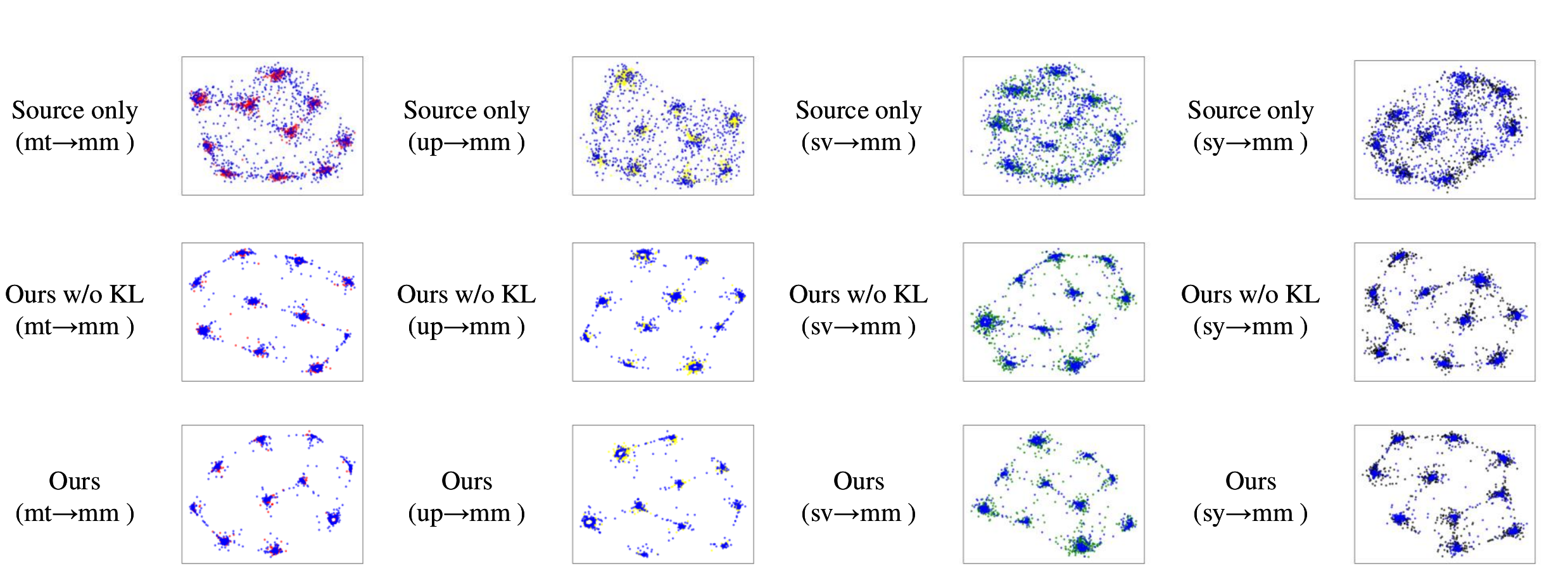}
\end{center}
   \caption{{The t-SNE visulization on Digit Recognition.} The red, yellow, green, black and blue points 
	are from domains \textbf{mt, up, sv, sy} and \textbf{mm} respectively. We used domains, \textbf{mt, up, sv, sy}, as source domains and \textbf{mm} as the target domain.} 
\label{fig:t-SNE-digit}
\end{figure*}

\begin{figure*}[t]
\vskip .1in	
\begin{center}
\includegraphics[width=0.70\linewidth]{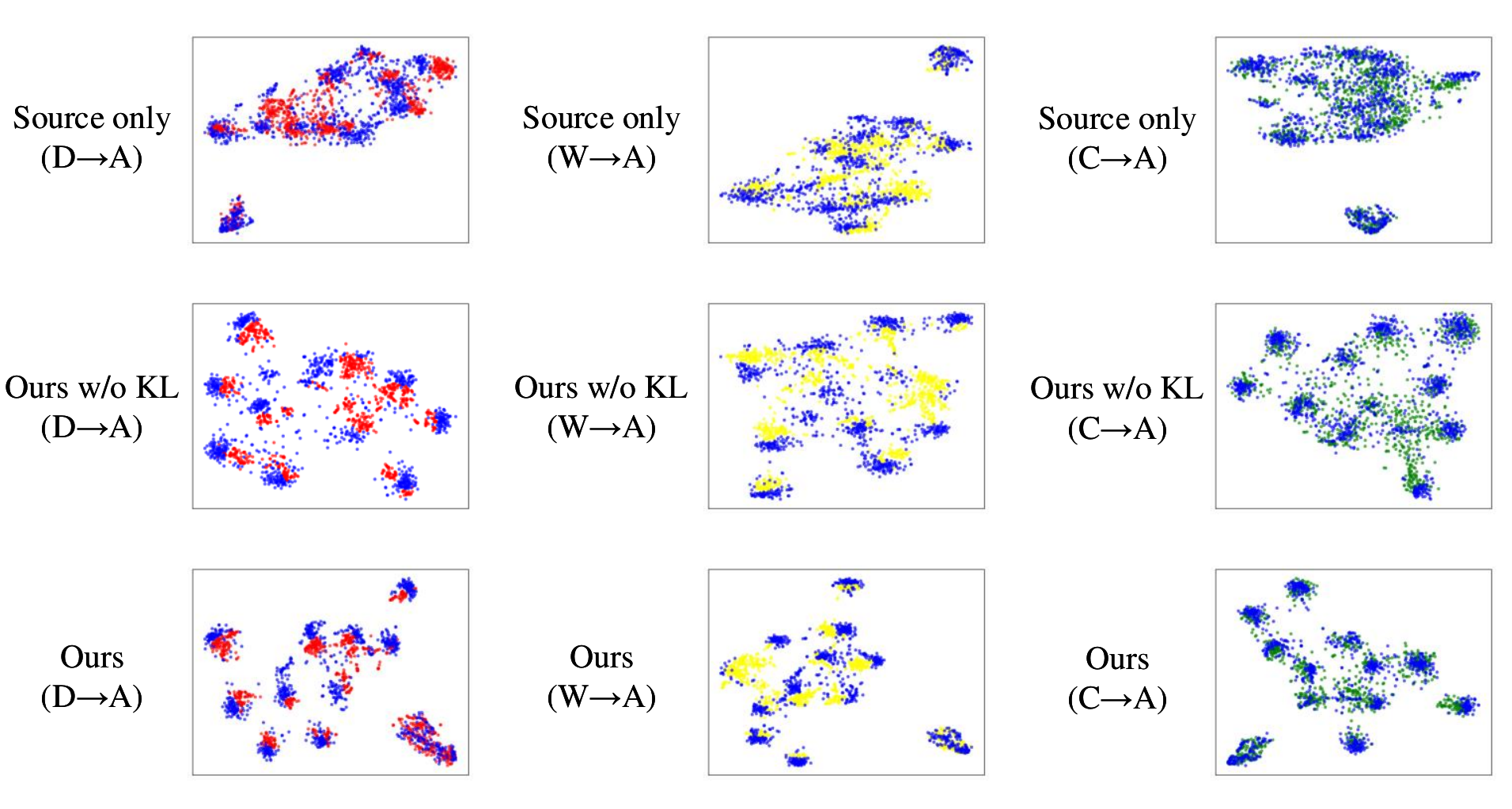}
\end{center}
   \caption{{The t-SNE visulization on Office-Caltech10.} The red, yellow, green and blue points represent data from domain \textbf{D, W, C, A} respectively. We use \textbf{D, W, C} as source domains and \textbf{A} as the target domain.}
\label{fig:t-SNE-office}
\end{figure*}

\subsection{Experiments on DomainNet}
DomainNet dataset is introduced in~\cite{peng2019moment}, which consists of six domains, namely \textbf{clp} ({\em Clipart}), \textbf{inf} ({\em Inforgraph}), \textbf{pnt} ({\em Painting}), \textbf{qdr} ({\em Quickdraw}), \textbf{skt} ({\em Sketch}) and \textbf{rel} ({\em Real}). 
Each domain has 345 classes of common objects. 
As shown in Table~\ref{table:DomainNet}, there are total 596,010 instances 
in the dataset and 1,728 instances per class. 
In our experiments, we chose 70\% from each domain for training and 30\% for testing. 
Benefiting from its large scale and wide
variety, the DomainNet dataset overcome 
the benchmark saturation issues of the state of the art domain adaptation datasets, which is of great significance to the study of domain adaptation.

We used the same comparison methods as in \cite{peng2019moment}, 
including \textit{DAN}~\cite{long2015learning}, \textit{RTN}~\cite{long2016unsupervised}, \textit{JAN}~\cite{long2017deep}, \textit{DANN}~\cite{ganin2014unsupervised}, \textit{ADDA}~\cite{tzeng2017adversarial}, 
\textit{SE}~\cite{french2017self}, \textit{DCTN}~\cite{xu2018deep} and \textit{MCD}~\cite{saito2018maximum}. 
Following the same setting as in \cite{peng2019moment}, we used AlexNet as the backbone for \textit{DAN}, \textit{
JAN}, \textit{DANN} and \textit{RTN}. 
We used ResNet-101 as the backbone for \(\it{M^3}\it{SDA}\), \textit{DCTN}, \textit{ADDA} and \textit{MCD}, 
while the backbone of \textit{SE} is ResNet-152. 
Same as \(\it{M^3}\it{SDA}\), our proposed method uses ResNet-101 as the backbone. 
In our method, the weights of {\em conv}1, {\em conv}2, {\em conv}3 and {\em conv}4 stages of all networks are shared.

The comparison results are reported in Table \ref{table:result-domainnet}.
From the table we can see that the average accuracy of our proposed method over the six multi-source domain adaptation tasks
is 44.3\% , which is 1.7\% higher than the best result produced 
by the comparison methods. 
Moreover, it is worth noting that on the task of \textbf{clp,inf,pnt,rel,skt}\(\to\)\textbf{qdr},
our proposed method outperforms other MSDA methods and single-source DA methods with notable performance gains. 
The work of \cite{peng2019moment} explains that the reason the multi-source methods perform poor
on this task is due to negative transfer~\cite{pan2009survey}. 
This suggests our proposed method can alleviate the problem of negative transfer.

\subsection{Further Analysis} 
\noindent{\bf Feature Visualization.} 
In our experiments on Digit Recognition and Office-Caltech10, we visualized the feature distributions
produced by 
the proposed ML-MSDA method to  validate its efficacy.
For comparison, we also visualized the results of the source-only baseline method,
and a variant ML-MSDA: ML-MSDA without JS-divergence (via KL-divergence). 

For easy observation, we show the distribution of each source domain separately together with the target domain. 
Fig. \ref{fig:t-SNE-digit} and Fig. \ref{fig:t-SNE-office} show the t-SNE~\cite{maaten2008visualizing} visulization of \textbf{mt, up, sv, sy}\(\to\)\textbf{mm} and \textbf{D, W, C}\(\to\)\textbf{A} respectively. 
We can see that for the proposed full approach the points of the target domain are closely centered
around the clusters of the source domains. 
This suggests our method can induce more transferable and discriminative features for the target domain.
\\

\noindent{\bf Ablation Study.} 
To further validate the efficacy of the proposed mutual learning network
and investigate the contribution of its different components, 
we conducted an ablation study to compare the proposed full approach ML-MSDA
with five of its variants: 
(1) ML-w/o condition-adv. This variant replaces the conditional adversarial feature alignment
with standard adversarial feature alignment by dropping the prediction probability vector ${\bf p}$
from $L_{adv}$.
(2) ML-w/o $L_E$. This variant drops the unsupervised entropy loss $L_E$ from ML-MSDA.
(3) ML-w/o $L_M$. This variant drops the prediction inconsisteny loss term $L_M$, the mutual learning term, from ML-MSDA.
(4) ML-guidance-inf. This variant performs training in the same way as ML-MSDA, but uses only the guidance network for inference in the testing phase.
(5) ML-branch-average-inf. This variant performs training in the same way as ML-MSDA, 
but drops the guidance network and uses the average of the branch networks for inference in the testing phase.
The comparison is conducted on four of the previously used multi-source domain adaptation tasks
and the results are reported in 
Table \ref{table:ablation}. 
We can see all the variants produced inferior results compared with the full ML-MSDA,
which suggests the components investigated, such as the entropy loss, conditional adversary, mutual learning regularization,
and the ensemble inference, are non-trivial for the proposed approach.
In particular, the variant ML-w/o $L_M$ leads to remarkable performance degradation,
which suggests 
the mutual learning regularization term $L_M$ is very important for the proposed ML-MSDA.
Moreover, the results also shows that it is beneficial to use both the guidance network and branch networks even in the testing phase.

\section{Conclusion}
In this paper we proposed a novel mutual learning network, ML-MSDA, for multi-source unsupervised domain adaptation.
It builds one adversarial adaptation branch network for each source-target domain pair
and a guidance adversarial adaptation network for the combined multi-source--target domain pair.
Mutual learning strategy is deployed to train these subnetworks simultaneously
by enforcing prediction consistency between the branch networks and the guidance network in the target domain.
We conducted experiments on a number of benchmark datasets.
The proposed ML-MSDA demonstrated superior performance than the state-of-the-art comparison methods.

\bibliographystyle{ieee}
\bibliography{egbib}
\end{document}